\documentclass[final,5p,times,twocolumn]{elsarticle}

\usepackage{lineno}
\usepackage{graphicx}
\usepackage{amssymb}
\usepackage{pdflscape}
\usepackage[english]{babel}
\usepackage[dvipsnames,svgnames,x11names]{xcolor}
\usepackage{booktabs,tabularx}
\usepackage{multirow}
\usepackage{hyperref}
\usepackage{amsmath}
\usepackage{commath}
\usepackage[english]{babel}
\usepackage[ruled]{algorithm2e}
\usepackage{caption}
\usepackage{subcaption}
\usepackage{balance}

\SetEndCharOfAlgoLine{}
\usepackage[leftmargin=6em,rightmargin=12em,indentfirst=false]{quoting}
\usepackage{siunitx}

\usepackage[final]{changes}
\usepackage{float}

\usepackage{lineno}
\usepackage{tikz}
\usepackage[export]{adjustbox}

\usepackage{graphicx}
\usepackage[utf8]{inputenc}
\usepackage[export]{adjustbox}
\usepackage{wrapfig}
\usepackage{dcolumn}

\makeatletter
\newcolumntype{T}[3]{>{\textfont0=\the@{#1}{#2}{#3}}c<{\DC@end}}
\makeatother

\usepackage{pgfplots}
\pgfplotsset{width=10cm,compat=1.9}

\usepackage{array}

\newcolumntype{L}[1]{>{\raggedright\let\newline\\\arraybackslash\hspace{0pt}}m{#1}}
\newcolumntype{C}[1]{>{\centering\let\newline\\\arraybackslash\hspace{0pt}}m{#1}}
\newcolumntype{R}[1]{>{\raggedleft\let\newline\\\arraybackslash\hspace{0pt}}m{#1}}

\usepackage{subfiles}

\usepackage{todonotes}

\RestyleAlgo{ruled}


\journal{Energy and Buildings}
\begin{document}
	
\begin{frontmatter}

\title{ALDI++: Automatic and parameter-less discord and outlier detection for building energy load profiles}

\author{Matias Quintana\,$^{1}$, Till Stoeckmann\,$^{2}$, June Young Park\,$^{3}$, Marian Turowski\,$^{2}$, Veit Hagenmeyer\,$^{2}$,  Clayton Miller$^{1, *}$}

\address{$^{1}$Department of the Built Environment, College of Design and Engineering, National University of Singapore (NUS), Singapore}
\address{$^{2}$Institute for Automation and Applied Informatics, Karlsruhe Institute of Technology, Germany}
\address{$^{3}$Department of Civil Engineering, University of Texas at Arlington, USA}
\address{$^*$Corresponding Author: clayton@nus.edus.sg, +65 81602452}

\begin{abstract}
Data-driven building energy prediction is an integral part of the process for measurement and verification, building benchmarking, and building-to-grid interaction.
The ASHRAE Great Energy Predictor III (GEPIII) machine learning competition used an extensive meter data set to crowdsource the most accurate machine learning workflow for whole building energy prediction.
A significant component of the winning solutions was the pre-processing phase to remove anomalous training data.
Contemporary pre-processing methods focus on filtering statistical threshold values or deep learning methods requiring training data and multiple hyper-parameters.
A recent method named ALDI (Automated Load profile Discord Identification) managed to identify these discords using matrix profile, but the technique still requires user-defined parameters.
We develop ALDI++, a method based on the previous work that bypasses user-defined parameters and takes advantage of discord similarity.
We evaluate ALDI++ against a statistical threshold, variational auto-encoder, and the original ALDI as baselines in classifying discords and energy forecasting scenarios.
Our results demonstrate that while the classification performance improvement over the original method is marginal, ALDI++ helps achieve the best forecasting error improving 6\% over the winning's team approach with six times less computation time.
\end{abstract}


\begin{keyword}

Smart meter \sep Load profile \sep Matrix profile \sep Discord detection \sep Portfolio analysis

\end{keyword}
\end{frontmatter}


\section{Introduction}
While indoor comfort is one of the main priorities of the built environment industry, this problem is linked with the concern of reducing the impact on global energy consumption and greenhouse gases emissions, which are 40\% and 30\%, respectively ~\cite{UnitedNationsEnvironmentProgramme2009}.
As smart meters become more prevalent, the availability of energy consumption data with a high resolution is rising~\cite{Park2018}.
These data are a catalyst for several applications that leverage machine learning-driven energy prediction methods~\cite{Amasyali2018-wj, Bourdeau2019-xt, Sun2020-jd}.
Prediction is a major component of urban energy optimization~\cite{Nutkiewicz2018-lw}, of retrofit scenarios~\cite{Grillone2020-fm,Seyedzadeh2019-hj}, a large building portfolio management~\cite{Park2019} and measurement and verification of energy savings implementations in both commercial~\cite{Granderson2016-wq,Granderson2017-lm} and residential buildings~\cite{Roth2021-be}. An array of machine learning methods have become popular in these applications, including neural networks~\cite{Rahman2018-zl,Tian2019-rb} and decision tree-based solutions such as Random Forests~\cite{Wang2018-ag,Ahmad2017-jz} and Gradient Boosting Machines~\cite{Touzani2018-sa}. 

\subsection{Data-driven building energy prediction and pre-processing importance}

Despite all the research momentum, there has been a lack of understanding of which steps in the machine learning process are most important for solution accuracy and usability. 
The ASHRAE Great Energy Predictor III (GEPIII) competition was held in 2019 as a way to address this lack of comparison issue~\cite{Miller2020b}. This competition had an extensive data set of whole building smart meters, and over 3,900 teams were competing to create the most accurate machine learning workflow. The winning solutions of this competition showed that ensembles of gradient boosting machines performed the best on a large data set.

An important insight from the GEPIII competition was that pre-processing of training data was a key differentiator amongst the top winning teams~\cite{Miller2020b}. 
This pre-processing step focuses on cleaning the energy consumption data from discords, data points, or sections that are not typical for the observed object.
When it comes to electricity price forecasting, statistical models such a threshold values and moving average filters~\cite{Afanasyev2019} or three times above the standard deviation~\cite{Cartea2005} are often used.
Pre-processing of training data for building energy prediction is especially necessary when there is a significant amount of anomalous signals in the data caused by non-routine behavior.
An error analysis from GEPIII captures the types of buildings that had more or less deviation from the predictions of the top 50 competitors~\cite{Miller2021-bf}.
This analysis showed that the meter types of hot water and steam as well as building use types such as \emph{Technology} or \emph{Food Service} were especially difficult to predict and therefore could benefit the most from enhanced pre-processing. 
This study also found that the electricity meter types and building use types such as \emph{Public Services} were easier to predict, therefore pre-processing may not be quite as important~\cite{Miller2021-bf}.

In the context of daily forecasts for energy consumption, only past typical daily profiles of a building should be taken into account to train the prediction model.
All other daily load profiles can be discord or anomalous clusters of behavior that could make machine learning models less generally accurate if used in the training data.

For this reason, it is essential to implement robust discord detection methods to generate "ready-to-use" data for further applications. 
The primary approach is checking the time series through visual analysis or by using IQR-based statistical outlier filtering~\cite{Himeur2021-ul,Cook2020-ye}. 
These methods are time-consuming and not practical for processing a large number of time series.
Another popular and classic technique is Symbolic Aggregate Approximation (SAX)~\cite{Lin2003} for energy consumption data~\cite{Miller20151, Capozzoli2018-lj} by looking into a symbolic representation of the time series based using letters of the alphabet to depict unique patterns.
Although this method has been used successfully~\cite{Miller20151} due to its easy interpretability, the symbols for the chosen segments represented by SAX are comprised of mean values which often miss important information like trends of extreme features of the segments.
Additionally, in addition to the window or segment size, the user would also require to choose the alphabet size, i.e., the number of discrete symbols to be used for identifiable patterns.

Furthermore, researchers rely on more complex ensemble~\cite{Araya2017-xq} or deep learning-based models~\cite{Liu2020-vk, Tang2014-jj}. 
For example, generative models such as auto-encoders are often used to learn the similarities of non-discord days such that a discord day is more accessible to detect~\cite{Araya2017}. 
The work done in~\cite{Fan2018} explores a wide variety of different auto-encoders and examines how well they are suited for detecting anomalies, as well as an ensemble of them.
However, these techniques do not address the semantics of the energy consumption data and, on the other hand, partly require many parameters to be set. 
Recently, a method called Matrix Profile (MP)~\cite{Yeh2017} has been used in energy consumption data to identify dominant building usage types~\cite{Nichiforov2020}. 
While this technique can locate discords in a more reliable way than simpler statistical models, it often requires significant parameter tuning, introducing different models, such as genetic algorithms, to optimize them~\cite{Fonseca2017a} properly.
The same technique is later introduced in~\cite{Park2020} as the building block for daily discord detection in a building energy data portfolio. Despite the momentum in anomaly and outlier detection, the implementation of such techniques still has several critical challenges related to the scalability of processes and ease-of-implementation \cite{Himeur2021-ul}. 

\subsection{ALDI++: Automating outlier and discord detection without parameter tuning}
This paper builds upon an existing daily discord detection algorithm (ALDI) presented in ~\cite{Park2020}.
The proposed framework, named ALDI++, bypasses the need for any user-defined parameter, and its performance is evaluated using the largest publicly available building energy meter dataset, the Building Data Genome 2 Project (BDG2)~\cite{Miller2020c}.
Specifically, the subset of data chosen as a case study was recently used in a machine learning competition on the Kaggle platform, which includes human-curated discord labels generated by the winning team of the competition~\cite{Miller2020b}.
For a holistic comparison, ALDI++ is evaluated first as a daily energy load profile discord classifier, second as an automated pre-processing tool for improving long-term time-series forecasting performance, and lastly in terms of its computation speed.
Comparisons are made with traditional statistical methods for discord removal, state-of-the-art deep learning methods, and the original ALDI algorithm~\cite{Park2020}. 
This approach aims to reduce the implementation configurations and efforts of anomaly detection in real-world data sets while increasing accuracy and usability. 
The proposed technique focuses on the ability to implement the process without setting parameters or optimization steps that may reduce implementation.

\section{Methodology}
\label{sec:methodology}

\subsection{Framework foundation: Automated Load profile Discord Identification (ALDI)}
As described in \cite{Park2020}, the original Automated Load profile Discord Identification (ALDI) framework \cite{Park2020} consists of the following three steps: 1) Hourly electrical energy consumption time series data is converted into a matrix profile (MP) to discover similar daily energy load profiles. 
MP calculation is a state-of-the-art data mining technique used to search similarities within time series data without any learnable parameter, unlike deep learning approaches~\cite{Yeh2017}.
2) The previous calculations are applied to all buildings in the same urban site, e.g., buildings on the same campus or district/city, and the daily MP values are grouped by the typical day types, i.e., Monday - Sunday.
Then, the Kolmogorov-Smirnov (KS) test~\cite{Journal2017} is used to compare the similarity of each days' MP distributions against the typical days' MP distributions ($MP_{site}(Monday)$ ... $MP_{site}(Sunday)$).
This statistical test answers the questions whether two samples come from the same probability distribution.
The computation results in two values: A $D-value$ which indicates the Euclidean distance between the samples' probability distributions and a $p-value$ which indicates the confidence level at which said distance was calculated.
3) Finally, the user evaluates the hypothesis test results qualitatively for only the MPs below a specific $p-value$ to identify the discord types (system malfunctions or irregular consumption pattern) for each site.

One limitation of the original ALDI is that the end-user still needs to specify a parameter, $p-value$, that serves as a confidence threshold of what daily load profiles are discords days or not.
Although this provides statistical soundness, an average user may not readily determine it. 
Moreover, to explore the building energy data itself, such parameter selection is uncertain and requires numerous interactive processes to calculate the results.
As a result, the authors in \cite{Park2020} recognize this and encourage further research into finding a systematic way of determining the parameter $p-value$ or circumventing it during the discord and non-discord day calculations.
Furthermore, a qualitative inspection of the discords days labeled by ALDI was needed in order to investigate their type of discord due to the unavailability of discord labels on the portfolios used.

\subsection{Unsupervised and parameter-less load profile discord detection (ALDI++)}
To bypass the need for a user-defined fix parameter such as $p-value$, our proposed extension ALDI++ repeats the first two steps of ALDI. Step 3 focuses on the distribution of the KS test results for each weekday regardless of their $p-values$, meaning unlike the original ALDI, no fixed $p-value$ is used as threshold.
Distances resulting from the KS test are bounded in $[0, 1]$ and indicate the Euclidean distance between the two cumulative distributions of the two samples; the lower the distance ($D-value$), the more similar their distributions are. 
This approach still requires a fixed $D- value$ as threshold such that $D < D_{threshold}$ are considered discords and vice-versa.
Thus, to group similar $D-values$, a univariate Gaussian Mixture Model (GMM) fits into the site's $D-values$ distribution such that specific Gaussian components are treated as discords.
A GMM is a parametric probability density function consisting of the weighted composition of several Gaussian normal distribution, a mixture model~\cite{Reynolds2015}.
To determine the parameters of a GMM, the iterative Expectation-Maximization (EM) algorithm is used which, based on an initial estimation of the GMM parameters, $\mu, \sigma, w$, and number of Gaussians, are updated iteratively until a termination criteria is met.
The parameters $\mu, \sigma$, and $w$, correspond to the mean, standard deviation, and weight for each normal Gaussian of the GMM.

\begin{figure}
	\centering
	\includegraphics[width=\linewidth]{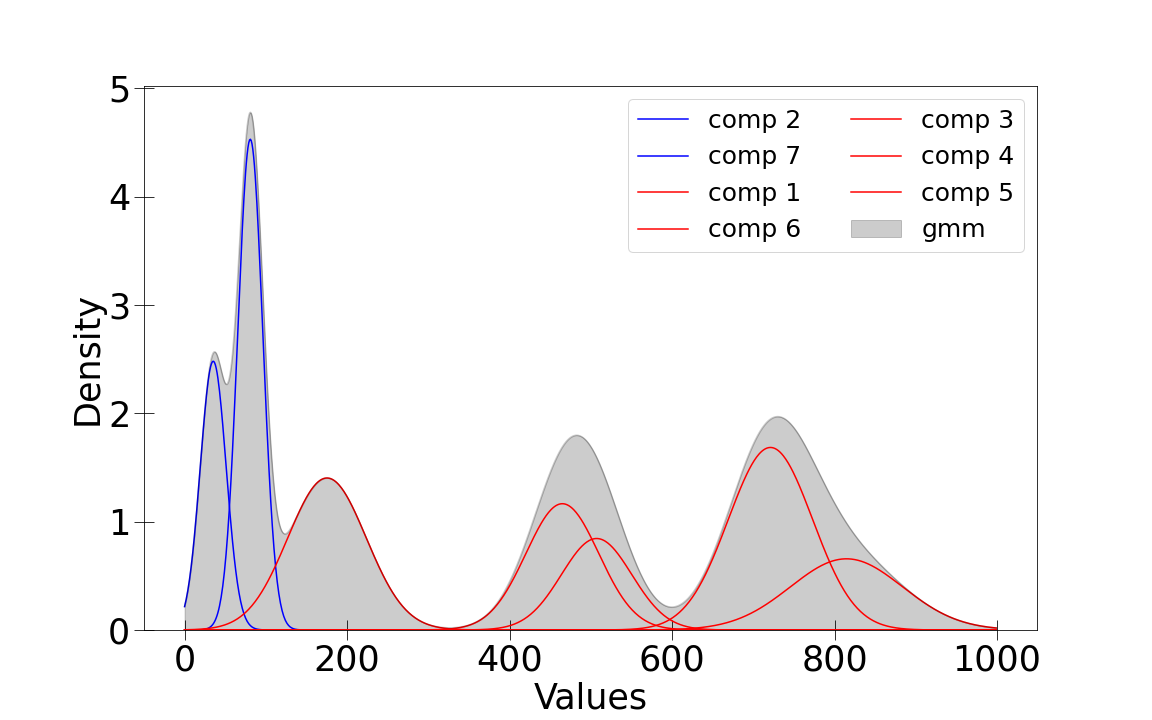}
	\caption{
	Example of one-dimensional data and a Gaussian Mixture Model (GMM) of 7 components fitted into it.
	The X-axis depicts the actual numerical values of the samples and Y-axis depicts the density or how many samples fall at said X value.
	The highest component's mean is 800, or 0.8, which means $k = 1 - 0.8 = 20\%$ and $20\% \times 7 = 1.4 \approx 2$ left most components (blue) are considered as non-discord and the remaining ones (red) are discords.
	}
	\label{fig:gmm_threshold}
\end{figure}

\begin{figure}
	\centering
	\includegraphics[width=\linewidth]{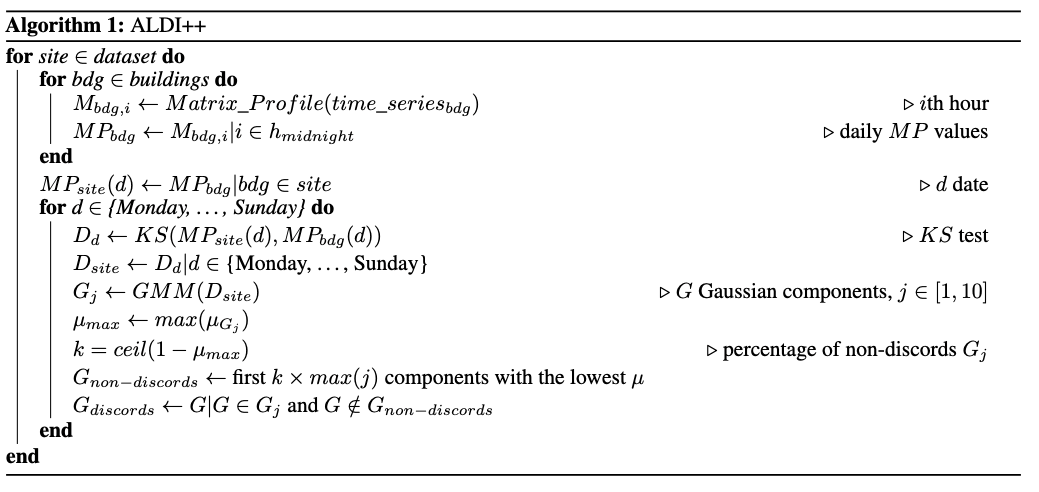}
	\label{alg:aldi_pp}
\end{figure}

For this approach, the number of Gaussians that are treated as discords are chosen dynamically based on the highest mean ($\mu_{max}$) of a component.
Any $\mu$ close to 1 means that the respective Gaussian component has high $D-values$, which translates to MP values further away from the respective weekday MP distribution and are potentially discords, with $\mu_{max}$ being a Gaussian components with the highest dissimilar distributions.
Nevertheless, discords in a dataset are expected to be a small number of samples~\cite{Prasad2009, Lazarevic2005}.
Therefore, instead of choosing a fixed threshold (e.g., top two components with the highest mean) the number of Gaussian components as non-discords, $k$ (components with higher $\mu$) is calculated as a percentage of the total number of components.
This percentage $k$ is calculated with the following Equation: $k = 1 - \mu_{max}$, rounded to the highest integer.
For example, if $\mu_{max} = 0.8$, the number of Gaussian components that are non-discord is $k = 1 - 0.8 = 0.2$ or 20\% of the total number of components.
If the GMM uses seven components, the first two Gaussian with the lowest $\mu$ are treated as non-discord Gaussians and the remaining five as discords.
While this may seem as ALDI++ over-classifying samples as discords since it treats more Gaussians as discords rather than non-discords, as the value of $\mu$ increases, the density of the Gaussian components decreases.
On the other hand, the components with the lowest $\mu$ tend to be clustered together with the highest density.
Figure \ref{fig:gmm_threshold} illustrates this example.
From here on, the $D-value$ of a given daily MP is evaluated on the GMM, and the discord/non-discord label is assigned depending on the Gaussian component it belongs. 
Algorithm \ref{alg:aldi_pp} shows the pseudo-code of the proposed framework.

\subsection{Evaluation}\label{sec:evaluation}

\subsubsection{Case study data source: Building Data Genome 2 (BDG2)}
\begin{figure}
	\centering
	\includegraphics[width=\linewidth]{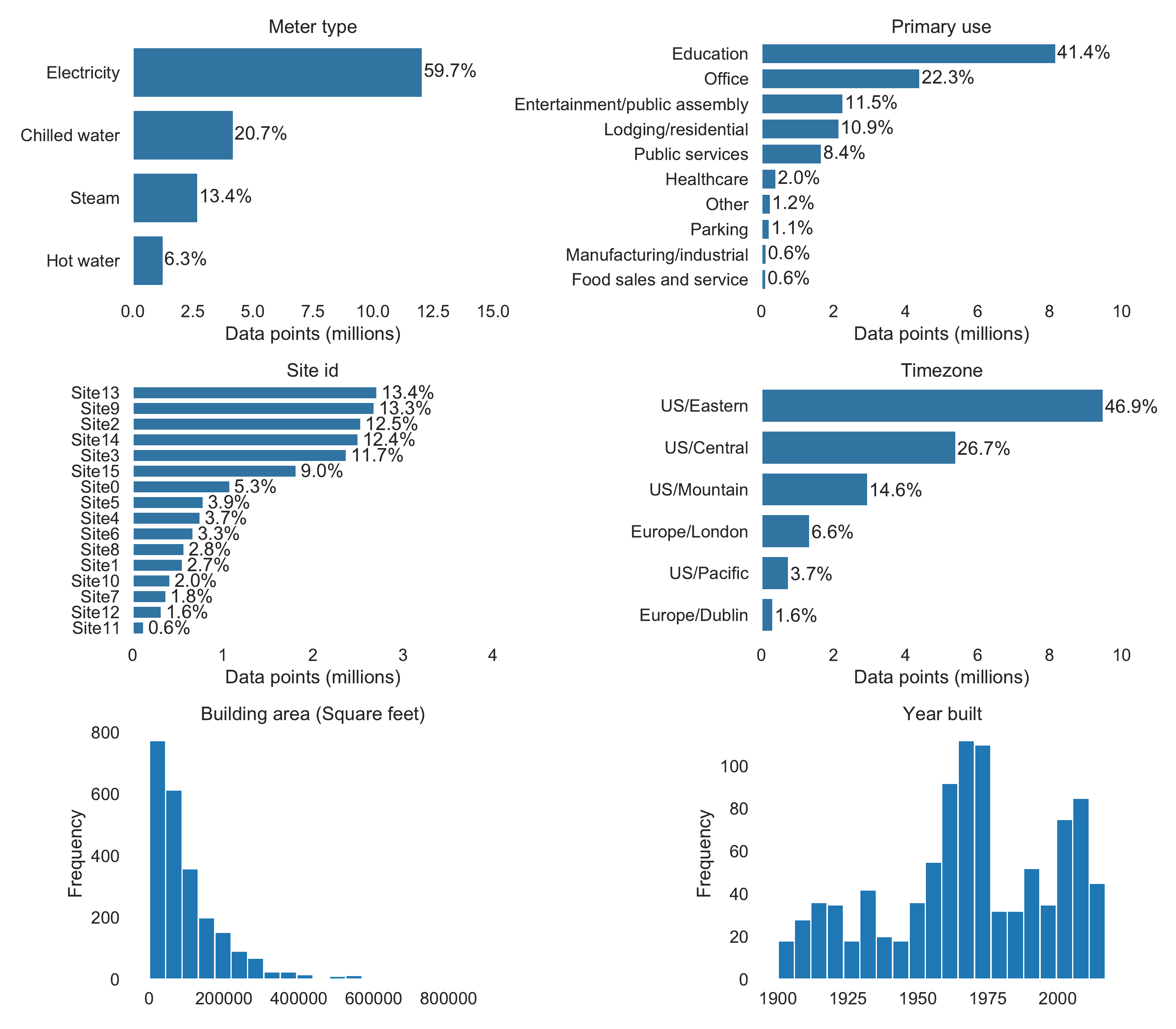}
	\caption{
	Overview of the BDG2 subset curated for the GEPIII competition (clockwise from upper left): Type of energy being measured (upper left); 
	The primary use type of the buildings in the dataset (upper right); 
	The time zone in which the buildings are located (middle right); 
	The year the buildings were constructed (lower right); 
	The gross floor area in sq. ft. of the buildings (lower left); 
	A breakdown of the amount of data collected from the 16 sites (middle left).
	}
	\label{fig:bdg2_overview}
\end{figure}

To evaluate ALDI++, we choose the most extensive publicly available building energy meter dataset, the Building Data Genome 2.0 (BDG2)~\cite{Miller2020c}
This dataset contains hourly electricity consumption data from 2016 to 2018 from 1,636 buildings across 19 North America and Europe locations.
A subset of this dataset was recently used in a machine learning competition on the Kaggle platform named the Great Energy Predictor III (GEPIII) competition\footnote{https://www.kaggle.com/c/ashrae-energy-prediction}~\cite{Miller2020b}.
The competition used the first year of data, 2016, as training data and then had the participants predict the entire subsequent two years.
The solutions of the top 5 winning teams are publicly available for reproducibility\footnote{https://github.com/buds-lab/ashrae-great-energy-predictor-3-solution-analysis}.
Figure \ref{fig:bdg2_overview} gives an overview of the data used for the GEPIII competition and this work.
Although four different streams of measurements are collected, this work uses only electricity measurements.
Only 16 sites, accompanied with their respective weather data, were used in the GEPIII, and the majority are universities; therefore, the most common building primary use type is education.
Almost three-quarters (73\%) of the data comes from buildings on university campus sites, and the remaining (27\%) come from city-wide municipal and healthcare building repositories.
As indicated in ~\cite{Miller2020b}, minimal data cleaning and processing were conducted on the data as the technical committee wanted conditions for the competitors to be as close to a real-world scenario in which data cleaning and pre-processing is an integral component of the winning contestants' solutions, which was ultimately highlighted by the discord detection strategy by the winning team.
In-depth details about the dataset and the contestants' solutions are provided by the competition organizers\footnote{https://github.com/buds-lab/ashrae-great-energy-predictor-3-overview-analysis}.

\subsubsection{Discord classification performance}
\label{discord_class_perf}
The winning team of the GEPIII competition manually curated an hourly discord/non-discord file for the entire training data of the competition.
This file was the result of approximately 8 hours of manual work inspecting all the portfolios time-series~\cite{Miller2020b} and will be used as \textit{ground-truth} discord/non-discord labels for this work.

There is no guarantee that the data points labeled as discords are indeed ground-truth labels and viceversa.
According to the winning team, a labeled discord is either a constant value for multiple timestamps, significant positive/negative outliers, or was picked up as anomalous during visual inspection of the energy consumption data by one of the team members.
Nevertheless, considering that the pre-processing step is what ultimately set apart the top 5 teams \cite{Miller2020b}, these discord labels are used as pseudo-ground-truth throughout this work.

\subsubsection{Impact on forecasting performance}
The main objective of the GEPIII competition was to find the most accurate (lowest forecasting error) modeling solution for long-term chilled water, hot water, and electricity forecasting.
While many teams used an ensemble of models, the most commonly used model was the LightGBM framework~\cite{Ke2017} due to its fast training speed, low memory usage, and high accuracy.
The modeling pipeline in this work, i.e., data pre-processing, feature engineering, and model training, are taken directly from the winning team's  solution\footnote{https://github.com/buds-lab/ashrae-great-energy-predictor-3-solution-analysis/tree/master/solutions/rank-1}.
Specifically, as part of the pre-processing step, all data points labeled as discords are removed from the training set and
the forecasting model chosen is the meter-specific LightGBM module\footnote{https://github.com/buds-lab/ashrae-great-energy-predictor-3-solution-analysis/blob/master/solutions/rank-1/meter\_split\_model/train.py} designed with only the electricity consumption as the forecasting variable.

\subsubsection{Benchmarking models}
\label{sec:benchmarking}
A total of three other discord detection models are evaluated for comparison: A 2-standard deviation model as a simple statistical model that considers all data points $\pm$2 standard deviation concerning the mean as discords, the original ALDI~\cite{Park2020} model as a baseline for the changes proposed by ALDI++, and a Variational Auto-encoder (VAE) as a commonly used deep-learning model, as shown in~\cite{Prasad2009, Araya2017, Fan2018}, combined with latent space clustering-based discord detection where discords belong to small clusters~\cite{Prasad2009}.
All models operate on a per-site basis, which means that a total of 16 instances for each model are used to generate the respective daily discord labels of each site.
Table \ref{tab:benchmarking} summarizes these models, alongside ALDI++, their characteristics and chosen parameters.

Although the VAE model is unsupervised, it requires training data to update its parameters; details about its hyper-parameters and architecture are listed in Table \ref{tab:vae_details}.
Thus, for each site, the data is divided into two halves: one half is used as train data and the remaining half as test data.
The discords days are labeled as the days which are members of the smallest of two clusters in the latent space.
This process of training on one half and testing on the other is repeated twice such that each half of the site's data is used for both train and test.

\begin{table}
	\centering
	\caption{Hyper-parameters of the Variational Auto-Encoder (VAE) model?}
	\begin{tabular}{cc}
		\toprule
		Hyperparameter & Chosen Value\\
		\midrule
		Num. of layers & 3\\
		Num. hidden units & 100, 50, 25\\
		Latent dimension size & 6\\
		Activation function & Tanh\\
		Learning rate & 0.0001 \\
		Epochs & 100\\
		Num. clusters & 2\\
		Clustering algorithm & K-means\\
		\bottomrule
	\end{tabular}
	\label{tab:vae_details}
\end{table}

\begin{table*}[]
    \centering
    \caption{
    Comparison of the benchmarking models regarding the number of parameters and the parameter values.
    The number of parameters considers both learned parameters (hyper-parameters) and parameters needed to be set by the user. 
    }
    \begin{tabular}{ccc}
        \toprule
        Model & Num. parameters & Parameters values\\
        \midrule
        2-Standard deviation (2SD) & None & -\\
        ALDI~\cite{Park2020} & 1 & $p-value= 0.01$\\
        Variational Auto-encoder (VAE) & High & Trained on train sets\\ 
        ALDI++ & None & -\\
        \bottomrule
    \end{tabular}
    \label{tab:benchmarking}
\end{table*}

As mentioned at the beginning of Section \ref{discord_class_perf}, the winning team discord labels will be used as ground-truth labels for evaluating the classification performance.
For the forecasting performance, the same modeling pipeline is used for all discord detection models where data points labeled as discords are removed from the training set.
This situation means that the main difference between discord detection approaches relies on the training set used and the quality and amount of training data points.
The forecasting performance using the winning team's discord labels is used as the baseline.
Our code is available for reproducibility in GitHub:\\ \texttt{https://github.com/buds-lab/aldiplusplus}.

\subsubsection{Quantitative metrics}
\label{sec:metrics}
As mentioned in previous subsubsections, different metrics are chosen to quantify the performance of ALDI++ and the benchmarking models.
Table \ref{tab:metrics} summarises the metrics.

\begin{table*}[]
    \centering
    \caption{
    Metrics used for the different evaluation scenarios and their characteristics.
    }
    \begin{tabular}{cccc}
        \toprule
        Metric & Range & Use case & Interpretation\\
        \midrule
        ROC-AUC & $[0, 1]$ & Discord classification & The higher the better\\
        RMSLE & $[0, +\infty]$ & Forecasting performance & The lower the better\\
        Computation time (minutes) & $[0, +\infty]$ & Discord label generation & The lower the better\\
        \bottomrule
    \end{tabular}
    \label{tab:metrics}
\end{table*}

\begin{figure}[h]%
    \begin{center} {
        \includegraphics[width=\linewidth]{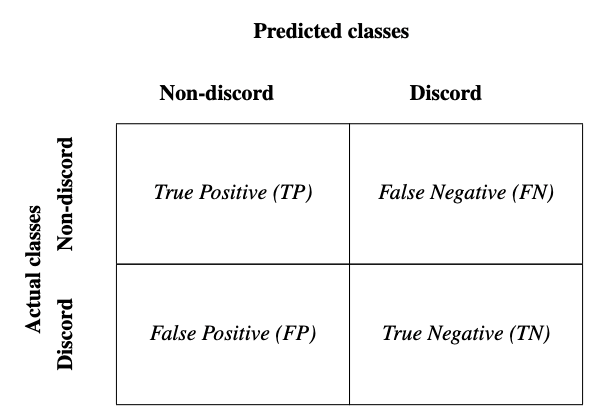}
        
        
        
        
        
        
        }
    \end{center}

    \caption{Confusion matrix for the discord and non-discord classification task}%
    \label{fig:confusion_matrix}%
\end{figure}

Firstly, to compare the labels generated by the GEPIII winning team against the discord labels that the models would produce, the Receiver Operating Characteristic Area Under the Curve (ROC-AUC) metric is chosen, which is based on values from the confusion matrix (Figure \ref{fig:confusion_matrix}).
In this way, the task becomes a binary classification problem where the winning team's labels are used as ground-truth, and the benchmarking models serve as classification models.
Given the nature of the resulting data where non-discords data points exceed the number of discord data points significantly, ROC-AUC is considered as a robust metric~\cite{Kim2017}.

This metric is calculated by determining the area of the Receiver Operating Characteristic (ROC), which contrasts the true positive rates (TPR) (y-axis) and false-positive rates (FPR) (x-axis)~\cite{Fawcett2006}.
Equation \ref{eq:tpr} and \ref{eq:fpr} show their calculation.

\begin{equation}
    TPR = \frac{TP}{TP + FN}
    \label{eq:tpr}
\end{equation}

\begin{equation}
    FPR = \frac{FP}{FP + TN}
    \label{eq:fpr}
\end{equation}

The value of ROC-AUC is bounded to $[0, 1]$ and the higher is the better, where a value of $0.5$ suggests the model is as good as randomly choosing either of the alternatives.
Confusion matrices are also used to properly assess whether the models have a tendency to misclassify discord days as non-discords (false positives) or vice versa (false negatives).

Secondly, the computation time required to generate the discord labels is calculated.
All experiments are performed on the same workstation (more details in Section \ref{sec:results}), but in order to account for minor differences, the runtime is calculated by running each model 10 times and reporting the average, and rounding it to the closest integer. 

Thirdly, as done in the GEPIII competition, the long-term energy forecasting results are reported using the Root Mean Squared Logarithmic Error (RMSLE).
When this metrics is used as a distance between two time series, e.g., $a$ and $b$, the logarithm function ensures that considerable deviations between the two time series to be compared do not significantly influence the result.
Equation \ref{eq:rmsle} shows its calculation.
\begin{equation}
    RMSLE(a,b) = 
       \sqrt{ \frac{1}{n}
              \sum\limits_{i=1}^{n} 
                  \big(
                  \log (a_{i} + 1)
                  -
                  \log (b_{i} + 1) 
                  \big)^{2}
       }
\label{eq:rmsle}
\end{equation}  

\section{Results}
\label{sec:results}
The workstation used for the experiments of this work is an Amazon Web Services (AWS) Elastic Compute Cloud (EC2) with the following characteristics:
Instance Type - \texttt{g4dn.4xlarge} (16 vCPUs, 64 GB RAM, and 600 GB disk),
AMI - Deep Learning AMI (Ubuntu 18.04),
Conda environment - \texttt{tensorflow2\_p36}.
The GEPIII winning team recommends a workstation with the same characteristics on their public solution.
The workstation used to calculate the computation runtime was a MacBook Pro (15 inch, 2017) with an Intel Core i7 Quad Core (2.9 GHz), 16 GB of DDR3 RAM and a Radeon Pro 560 4GB graphics card.

\subsection{Classification performance}
\begin{figure*}
	\begin{subfigure}{0.49\textwidth}
    	\centering
    	\includegraphics[width=1\linewidth]{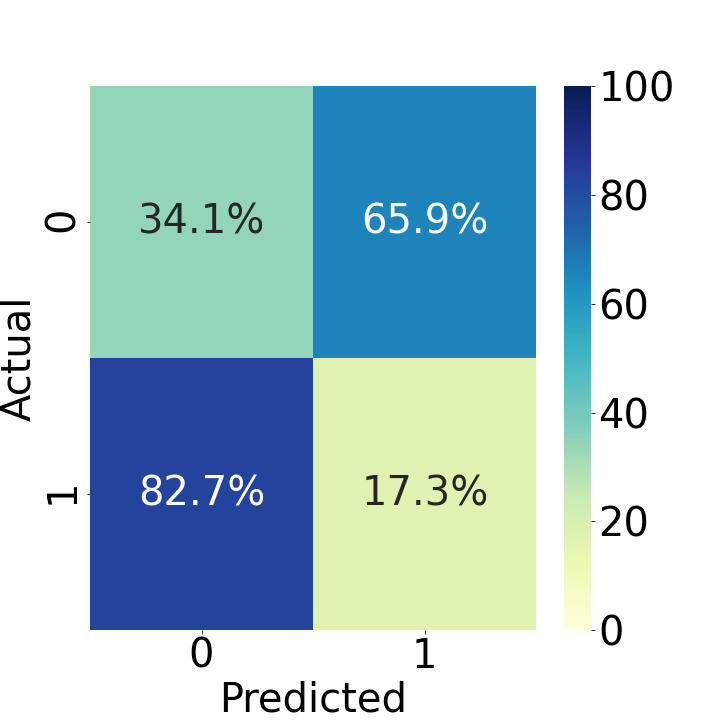}
    	\caption{2-standard deviation classification matrix, \\ROC-AUC: 0.2568}
    	\label{fig:roc_std}
	\end{subfigure}
	\hspace{.35cm}
	\begin{subfigure}{0.49\textwidth}
    	\centering
    	\includegraphics[width=1\linewidth]{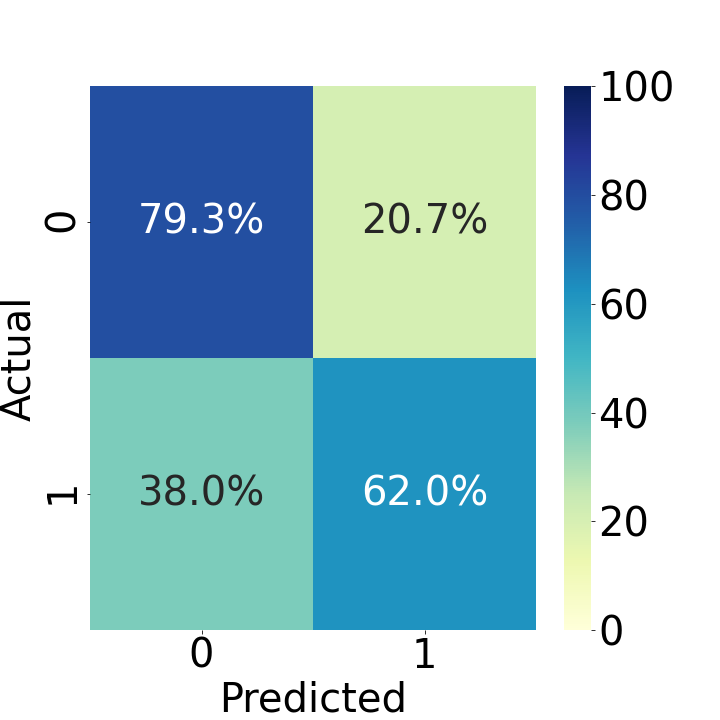}
    	\caption{ALDI~\cite{Park2020} classification matrix, \\ROC-AUC: 0.5795}
    	\label{fig:roc_aldi}
	\end{subfigure}
	~
	\begin{subfigure}{0.49\textwidth}
    	\centering
    	\includegraphics[width=1\linewidth]{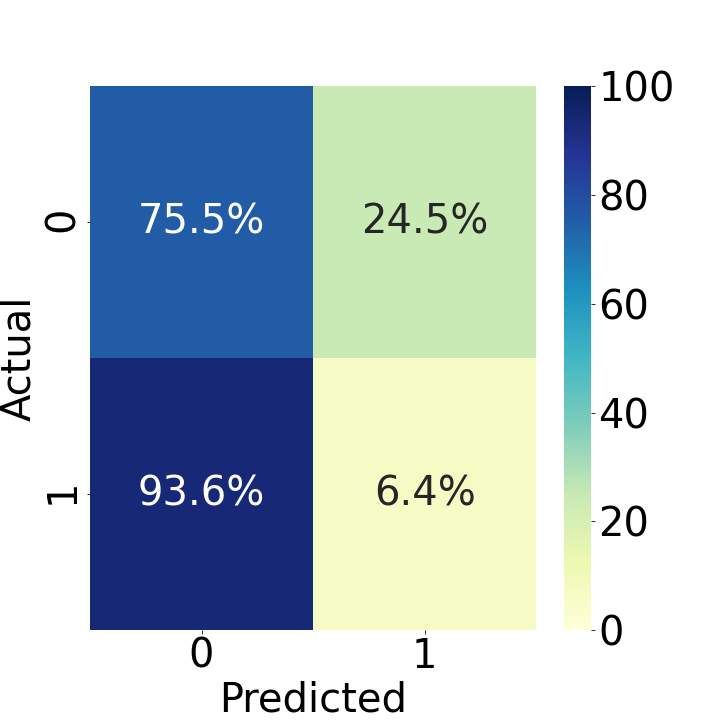}
    	\caption{VAE classification matrix, \\ROC-AUC: 0.4093}
    	\label{fig:roc_vae}
	\end{subfigure}
	\hspace{.35cm}
	\begin{subfigure}{0.49\textwidth}
    	\centering
    	\includegraphics[width=1\linewidth]{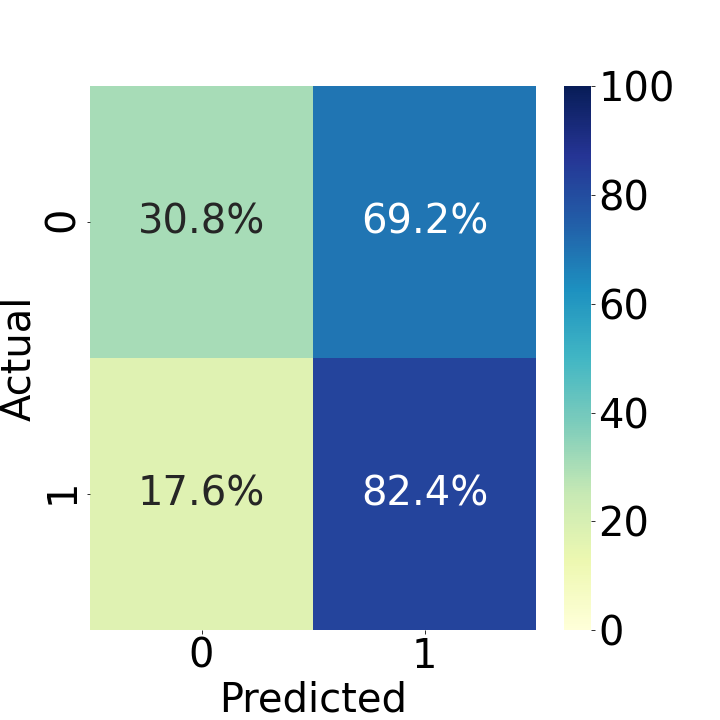}
    	\caption{ALDI++ classification matrix, \\ROC-AUC: 0.5818}
    	\label{fig:roc_aldipp}
	\end{subfigure}
	\caption{Classification matrices and ROC-AUC for all benchmarking methods: \ref{fig:roc_aldi} original ALDI~\cite{Park2020}, \ref{fig:roc_std} 2-Standard deviation (2SD) approach, \ref{fig:roc_vae} Variational Auto-Encoder (VAE), and \ref{fig:roc_aldipp}:ALDI++.
	Discord labels are represented by $1$ and non-discords labels as $0$.
	}
	\label{fig:roc_all}
\end{figure*}

The labels generated by the GEPIII winning team are given for each data point in the training set, meaning they are provided on an hourly basis; however, the approaches tested on this work generate discord labels on a daily profile basis.
Thus, the hourly discord labels used as reference are transformed into daily discord labels.
Within the pseudo-ground-truth labels, an entire day is considered a discord day if at least 14 hourly readings have been labeled as discord hours.
While 12 hourly readings could be enough to consider the entire day as discord, 14 hourly readings were chosen based on the work done in~\cite{Lazarevic2005, Prasad2009} which suggests that discords or outliers constitute 5-10\% of a dataset.
Accordingly, the hourly threshold is calculated by considering a minimum of half a day of hourly discords plus 10\% of potentially hourly outliers: $\frac{24}{2} + 24 \times 0.1 0 = 14.4 \sim 14$.
Figure \ref{fig:roc_all} shows the confusion matrices of all four benchmarked approaches against the pseudo-ground-truth labels and their respective ROC-AUC value.
Discord are represented by $1$ and non-discords by $0$.

ALDI and ALDI++ have a similar average performance slightly above $0.5$, which suggests a close to random classification performance.
The VAE and original ALDI~\cite{Park2020} are the best approaches in terms of correctly labeling non-discords days as non-discords, 75.5\% and 79.3\%, respectively. 
Still, the VAE severely underperforms in true negatives ratio (6.4\%), i.e., predicting it is a discord day when it truly is a discord day.
Conversely, ALDI++ overclassifies non-discords days as discords days, resulting in the highest false-positive ratio among all models (around $\sim$65\% compared to 20-24\%).
However, when compared to the original ALDI~\cite{Park2020}, the newly proposed method ALDI++ increases the discord day detection by 20.4\% (from 62\% to 82.4\%).

\subsection{Forecasting performance}
\begin{figure}
	\centering
	\includegraphics[width=\linewidth]{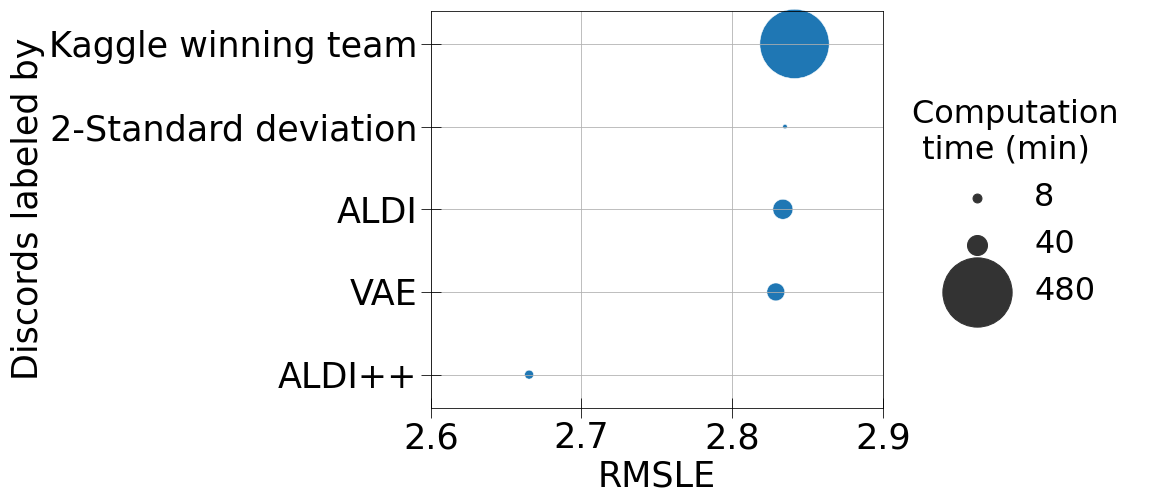}
	\caption{
	Forecasting RMSLE comparison on the GEPIII public test set.
	According to the respective discord labels, the only difference across the modeling pipeline for all methods is the discarded training data points.
	The size of the bubble is proportional to the computation time required to label the training dataset.
	}
	\label{fig:forecasting}
\end{figure}

Following the same modeling pipeline, one LightGBM forecasting model is trained for each discord detection model, with the only difference between them being the data pre-processing.
During this pre-processing, data points labeled as discords are removed from the train set.
Since the dataset used has hourly data points, the daily discord labels generated are converted to an hourly resolution, i.e., if a day is labeled as discord, all 24 hourly data points are marked as discords.
Figure \ref{fig:forecasting} shows a bubble plot with the RMSLE on the test set of the GEPIII competition where the size of bubbles is proportional to the computation time required for each benchmarked approach to generate all discord labels.

The ``Kaggle winning team'' approach uses the winning team's discord labels directly and is used as a baseline.
The RMSLE obtained in Figure \ref{fig:forecasting} differs from their winning RMSLE\footnote{https://www.kaggle.com/c/ashrae-energy-prediction/leaderboard}, 2.841 and 1.231 respectively, mainly because the modeling pipeline used in this work is a subset of their final ensemble approach, and it only uses the electricity data as input instead of the four available data streams.
While their exact final solution could have been replicated, the main goal of this work is to evaluate the impact of detecting potential discords days based on electricity meter data and removing them from the training set and not the modeling itself.

Overall, the benchmarked models 2-Standard deviation, ALDI, and VAE achieve a similar performance compared to the winning team performance, 2.835, 2.834, and 2.829 respectively, compared to the winning team's 2.841 score.
While these results are marginally worse than the winning team performance, the computation time for the calculation of the discord labels is reduced to one twelfth, from 480 minutes (8 hours) to 40 and 32 minutes for ALDI and VAE, respectively; whereas the 2-Standard deviation model only takes close to one minute of computation runtime.

Finally, the proposed method ALDI++ outperforms all models with an RMSLE of 2.665 and a computation time of 8 minutes.
Compared to the winning team performance, this represents an RMSLE improvement of 6\% and a computation time that is 60 times smaller.

\section{Discussion}
The proposed method, ALDI++, can label daily energy load profiles as discords similarly to current state-of-the-art methods and reduce the long-term forecasting error by removing these discords from the train data.
Specifically, it is determined that:
i) Using a human-labeled energy consumption dataset as a reference, all benchmarked models achieve a slightly above random discord classification ROC-AUC of at most $\sim$0.58;
ii) data-driven discord detection approaches reduce the forecasting error compared to human-labeled discords. They can generate the discord labels with at least 6 times less computation time.

\subsection{Discord classification}
As mentioned in Section \ref{sec:evaluation}, the validity of the human-labeled discord labels is hard to verify and at most serve as a pseudo-ground-truth reference for this work.
The classification ROC-AUC scores highlight this mismatch of what humans can identify as discords versus what the data-driven algorithms can detect.
The winning team, and in fact all the top three teams, used a combination of detecting constant values, visual identification, and domain-knowledge (i.e., some team members were Civil Engineers familiar with energy meter data) to label data points as discords as best as they could.
This process took them a considerable amount of time (8 hours) and was not automated, though, in return, it allowed them to win the competition~\cite{Miller2020b}.

Nevertheless, all benchmarked data-driven models achieve performance within 10\% of the performance of a random classifier (ROC-AUC of 0.5) with respect to these labeled data points.
Notably, the proposed method generates 38\% more discord labels than the winning team.
It remains an open question whether these false-positive discords labeled by the data-driven methods are discords that the winning team failed to identify or are truly discord data points in the data. 
By far, our human-generated discords labels are the most accurately identified ones by experts that we can use to evaluate ALDI++'s performance of discord detection.
In addition to our labels, we could add another curated discord or fault detection time-series dataset to assess the classification performance of these data-driven discord detectors.
We leave this validation for future work.

\subsection{Forecasting performance}
As seen in Figure \ref{fig:forecasting}, all data-driven models improve the forecasting performance compared to what is achievable using the human-generated labels, and the required time to generate these discords labels is substantially lower.
These results seem to contradict the poor classification performance mentioned in the previous subsection; however, we believe this is a limitation of how the pseudo-ground-truth labels were generated.
The proposed method, ALDI++, achieves the lowest RMSLE at one of the lowest time to calculate discords. However, it also identifies 44\% of the training data as discords; in comparison, the winning team labeled 6\% of the train data as discord.
We hypothesize that the reason behind the increase in forecasting performance with a much smaller training set is due to the quality of the non-discord data points used for training.
Inadvertently, what ALDI++ might be doing is not necessarily picking up all the truly discord days but filtering non-representative days do not benefit the forecasting model since it is implausible that 44\% of a dataset are outliers.
Even so, as domain knowledge in the discord labeling played a significant role in the winning team's solution, future works could leverage experts' annotations on what ALDI++ or other data-driven methods generate as potential discords.

\subsection{Practical applications}
A vital user group of this tool are facility managers and building operators.
These users are more interested in looking at the days at which discords occur and then investigating the potential reasons behind the malfunction.
For this purpose, the tool at hand should be easy to understand and should require fewer user-defined parameters.
While filtering out discord days with a simple statistical model like $\pm$2 standard deviation from the mean could immediately remove very high or low values, it overlooks patterns and trends.
Then, unlike the predecessor ALDI~\cite{Park2020}, ALDI++ does not require the user to manually set any parameter, although some engineering choices are internally made for the number of components in the GMM.
Moreover, compared to deep learning methods like the VAE, ALDI++ has the advantage of computing the results at the time of execution without any training or parameter updates, which could be more user-friendly for those not familiar with computing knowledge. 
The foundation of this advantage relies on matrix profile and its ability to accurately and quickly compute similarities within a time-series~\cite{Yeh2017}.

On the other hand, the application of discord detection to improve forecasting performance targets a particular audience of predominantly data scientists.
These users might not be interested in correctly narrowing down the days at which a discord or malfunction of a building occurs but rather prioritize the forecasting performance.
For this purpose, ALDI++ offers the best performance at the cheapest computing cost (Figure \ref{fig:forecasting}.
As shown in the GEPIII competition, among $3,614$ teams, the main differentiator among the top 3 winning teams was the pre-processing and data cleaning methods they used~\cite{Miller2020b}.
As data-driven modeling and machine learning libraries and frameworks become more widespread and available, people with relatively low technical backgrounds are able to train and test complex models; there is an increasing interest in the quality of the data rather than the model itself.
This new ``data-centric'' paradigm focuses more on the issues of data collection, quality, and cleaning, particularly for  real-world deployed systems~\cite{Paleyes2021}, which also becomes apparent through its first ever competition held in September 2021 called Data-Centric AI\footnote{https://worksheets.codalab.org/worksheets/0x7a8721f11e61436e93ac8f76da83f0e6}.
Aligned with this, ALDI++ is a computationally cheap and automatic tool for data-centric applications.

Finally, as mentioned in Section \ref{sec:benchmarking}, we publish our code on GitHub: \texttt{https://github.com/buds-lab/aldiplusplus}, to not only allow reproducibilty of our experiments and further benchmarking but to facilitate the use of ALDI++ by the building and data science community.

\section{Conclusion}
ALDI++ is a parameter-less discord detection method that builds on previous methods and improves forecasting performance at a fraction of the computation time needed by manual efforts.
We evaluated ALDI++ with the human-generated discords, which are the most accurately identified labels by the winning teams of GEPIII. 
Compared to other methods (ALDI~\cite{Park2020}), deep learning methods (VAE) and statistical methods, ALDI++, by removing labeled discords from the training set, achieves the best forecasting RMSLE, improving 6\% over a simplified version of a machine learning competition winning team's solution.
Also, it can generate these labels in eight minutes, five times faster than its predecessor and 60 times faster than the visual inspection of the machine learning competition winning team.

Unlike related literature on discord detection for energy consumption, we unravel the scenario where, for forecasting performance purposes, finding representative data points for training is more important than accurately finding all true discords.
Nevertheless, more need for accurately labeled discord data sets is required to assess the supervised learning classification capabilities of different models properly.
For the practical applications of ALDI++, our research could enable real-time detection of potential discord days, which could be more helpful in building facility managers and operators.
While this method is only tested on electricity consumption data, it can be extrapolated to other time-series portfolio problems in the built environment (e.g., water consumption, air quality, transportation demand).

\section*{CRediT author statement} 
\textbf{MQ:} Conceptualization, Methodology, Writing - Original Draft.
\textbf{TS:} Conceptualization, Methodology, Software, Validation, Formal analysis, Investigation, Visualization, Writing - Reviewing and Editing.
\textbf{JYP:} Conceptualization, Resources, Visualization, Supervision, Writing - Reviewing and Editing.
\textbf{MT:} Supervision, Writing - Reviewing and Editing.
\textbf{VH:} Supervision, Writing - Reviewing and Editing.
\textbf{CM:} Conceptualization, Resources, Visualization, Supervision, Project administration, Writing - Reviewing and Editing.

\section*{Acknowledgement}
This research was funded by the Republic of Singapore's National Research Foundation through a grant to the Berkeley Education Alliance for Research in Singapore (BEARS) for the Singapore-Berkeley Building Efficiency and Sustainability in the Tropics (SinBerBEST) Program. BEARS has been established by the University of California, Berkeley as a center for intellectual excellence in research and education in Singapore.
This work was also funded by the Helmholtz Association's Initiative and Networking Fund through Helmholtz AI and the Helmholtz Association under the Program ``Energy System Design''.
\bibliographystyle{unsrt}
\bibliography{references}

\end{document}